\documentclass{article}

\usepackage{custom_preprint}

\usepackage{xspace}
\usepackage[utf8]{inputenc} 
\usepackage[T1]{fontenc}    
\usepackage{hyperref}       
\usepackage{url}            
\usepackage{booktabs}       
\usepackage{amsfonts}       
\usepackage{nicefrac}       
\usepackage{microtype}      
\usepackage{xcolor}         
\usepackage{amsmath}
\usepackage{pgfplots}
\pgfplotsset{compat=1.18} 
\usepackage{tikz}
\usepackage{pgfplotstable}
\usepackage[capitalize,noabbrev]{cleveref}
\usepackage{hyperref}
\usepackage{pifont}
\newcommand{\cmark}{\ding{51}}%
\newcommand{\xmark}{\ding{55}}%
\newcommand{\xpm}[1]{{\tiny$\pm#1$}}

\usepackage{authblk}
\usetikzlibrary{arrows.meta, backgrounds, calc, fit, positioning}

\usepackage[textsize=tiny]{todonotes}

\def\methodName{PhenoKG}

\title{\methodName{}: Knowledge Graph-Driven Gene Discovery and Patient Insights from Phenotypes Alone}
\author[ 1,2]{Kamilia Zaripova
\thanks{Corresponding author: \texttt{kamilia.zaripova@tum.de}}
}
\author[1,2]{Ege Özsoy}
\author[1,2,3]{Nassir Navab}
\author[1,2]{Azade Farshad}

\affil[1]{Department of Computer Science, Technical University of Munich}
\affil[2]{Munich Center for Machine Learning (MCML), Munich, Germany}
\affil[3]{Johns Hopkins University, Baltimore, Maryland, USA}

\begin{document}

\maketitle

\begin{abstract}
Identifying causative genes from patient phenotypes remains a significant challenge in precision medicine, with important implications for the diagnosis and treatment of genetic disorders. We propose a novel graph-based approach for predicting causative genes from patient phenotypes, with or without an available list of candidate genes, by integrating a rare disease knowledge graph (KG). Our model, combining graph neural networks and transformers, achieves substantial improvements over the current state-of-the-art. On the real-world MyGene2 dataset, it attains a mean reciprocal rank (MRR) of 24.64\% and nDCG@100 of 33.64\%, surpassing the best baseline (SHEPHERD) at 19.02\% MRR and 30.54\% nDCG@100. We perform extensive ablation studies to validate the contribution of each model component. Notably, the approach generalizes to cases where only phenotypic data are available, addressing key challenges in clinical decision support when genomic information is incomplete.
\end{abstract}

\section{Introduction} \label{sec:into}
With over 300 million individuals affected globally, rare genetic disorders represent a profound challenge for healthcare systems, largely due to their low incidence and wide-ranging symptoms \citep{nguengang2020estimating}. Even with rapid developments in genomic science, the path to an accurate diagnosis is often long and convoluted, frequently requiring years of medical appointments and consultations with various specialists \citep{adams2024addressing}. This extended diagnostic process—often referred to as a "diagnostic odyssey"—is made more difficult by the limited number of experts in the field, the absence of unified diagnostic criteria, and the scarcity of annotated datasets necessary for training AI-driven diagnostic tools \citep{decherchi2021opportunities}. As a consequence, approximately 70\% of patients in search of a diagnosis remain without one, and the genetic causes of nearly half of all Mendelian conditions—those resulting from mutations in a single gene—remain unidentified \citep{gahl2012national, chong2015genetic}. These delays not only increase the likelihood of unnecessary or repetitive testing but also hinder timely intervention, often leading to preventable deterioration in patient health.

Seminal and recent studies on rare genetic diseases have shown that pinpointing the pathogenic variant in a gene responsible for a patient’s condition greatly improves diagnostic accuracy and informs downstream clinical management~\citep{boycott2013rare,posey2019insights}. Historically, clinicians manually integrated structured and unstructured clinical observations—spanning physical traits, symptom trajectories, and laboratory findings—with genomic data produced by exome or genome sequencing \citep{james2016visual}. Although many centres now employ semi‑automated decision‑support pipelines using methods for
identification of potential patients with rare conditions, like \citep{prakash2021rarebert, thompson2023large}, expert interpretation remains indispensable. The collection of observable characteristics (the phenotype) provides essential context for interpreting the genotype —the individual’s complete set of genetic variants — which often contains variants of uncertain significance.
When a pathogenic or likely pathogenic variant is identified in a gene whose gene‑disease relationship is rated Definitive or Strong and whose known disease spectrum fits the patient’s phenotype, that variant can be considered causative under ACMG/AMP and ClinGen criteria \citep{richards2015standards, strande2017evaluating}. This information is crucial not only for diagnosis but also for prognosis, genetic counseling, and, in selected cases, therapeutic decision‑making.

Despite advances in sequencing, variant‑effect prediction, and phenotype‑driven gene prioritisation, overall diagnostic yields for rare Mendelian disorders still hover around 30–50\%—and are lower in under‑represented ancestries and in atypical clinical presentations. These persistent gaps underscore the need for next‑generation machine‑learning models, graph‑based approaches, and multimodal frameworks that can integrate heterogeneous biomedical data at scale while preserving clinical interpretability and portability across diverse populations \citep{decherchi2021opportunities}.

Most diagnostic pipelines assume that clinicians have already narrowed the search to a hand‑curated list of ``candidate'' genes harboring patient‑specific DNA variants—an approach limited by current medical knowledge and expertise, and requiring substantial manual effort. We present a scalable alternative that starts from a patient’s phenotype terms, enriches them with a biomedical knowledge graph (KG), and ranks \(\approx 8000\) genes by their probability of being causative. This serves as either a direct prediction method or as a pre-filter to help clinicians avoid testing variants across the entire genome. We focus on Mendelian diseases with a single causative gene. Without a pre‑filtered list, the model achieves \(15.15 \pm 3.33\%\) accuracy (correct gene among the top-ranked suggestions). With an optional expert-curated variant (candidate) list (\(\approx 20\) genes), accuracy rises to \(83.96 \pm 1.45\%\). The constructed patient graph structure enables the model to suggest genes for the prioritization evaluation outside expert-curated lists, capturing patient-specific phenotype-disease-genotype relationships. In theory, this provides clinicians with an additional discovery layer; however, this requires prospective validation, as only follow‑up variant testing can confirm the predictions.

Our contributions are summarized as follows:
\begin{enumerate}
    \item We introduce PhenoKG, a method that captures patient-specific uniqueness via graph structures and gene representations, outperforming current state-of-the-art approaches.
    \item We demonstrate that our model performs well both with and without a candidate list through quantitative evaluation.
    \item Our experiments show that models trained solely for causative gene prediction underperform on patient-like retrieval tasks when evaluated by true gene alone. However, strong performance on the primary task suggests that patients grouped closely in embedding space may share common biological characteristics beyond a single gene, which warrants further investigation.
\end{enumerate}

\section{Related Works} \label{sec:into}
To address diagnostic odysseys, researchers have explored three main machine learning strategies: genotype-based methods, phenotype-based methods, and hybrid models. Genotype-based Methods leverage genomic sequencing data to identify variants associated with specific diseases. Techniques such as variant frequency analysis, pathogenicity prediction, and gene-disease databases are commonly used. Tools like MutationTaster \citep{steinhaus2021mutationtaster2021}, CADD \citep{rentzsch2019cadd}, and M-CAP, a clinical pathogenicity classifier \citep{jagadeesh2016m}, have shown utility but are limited by the need for comprehensive variant annotations and training data.  

Phenotype-driven tools prioritize diseases or genes by comparing a patient's phenotypic abnormalities to curated knowledge bases like Phenolyzer \citep{yang2015phenolyzer,kohler2009clinical} that use prior information to implicate genes involved in diseases and others \citep{jagadeesh2019phrank, peng2021cada, rao2018phenotype}. Facial phenotyping tools like DeepGestalt \citep{gurovich2019identifying}, GestaltMatcher \citep{hsieh2022gestaltmatcher}, PEDIA \citep{hsieh2019pedia}, and others \citep{duong2022neural,hong2021genetic, shukla2017deep} have demonstrated the potential to bridge phenotype-genotype associations, though they often struggle with complex or novel presentations.

Hybrid approaches combining genotype and phenotype data have improved diagnostic accuracy, including Bayesian frameworks integrating probabilities from multiple data sources \citep{robinson2020interpretable,javed2014phen} and deep learning models predicting disease or gene likelihood from phenotype-genotype associations \citep{smedley2015next,li2019xrare,birgmeier2020amelie,yoo2021inphernet,anderson2019personalised}. A recent example is AI-MARRVEL (AIM) \citep{mao2024ai}, a random forest that ranks variants genome-wide by combining patient variant-call files with Human Phenotype Ontology (HPO) terms. In contrast, \methodName\ operates without sequencing data, prioritizing genes rather than variants. Using only phenotype data and a knowledge graph, it serves as an upstream triage tool to reduce clinicians’ search space, accelerate gene prioritization, and surface plausible candidate genes beyond curated panels.

Other methods, such as Shepherd \citep{alsentzer2022few} and CADA \citep{peng2021cada}, leverage KGs to enrich patient information. CADA builds a phenotype–gene KG, learns Node2Vec embeddings, and ranks candidate genes by embedding similarity to patient HPO term vectors. Like our approach, CADA requires only phenotype data but underperforms compared to Shepherd due to KG design and model choices; thus, we benchmark only against Shepherd. Shepherd introduces three models with shared pre-training for disease prediction, gene prioritization, and ``patient-like-me'' prediction; we focus solely on gene prioritization. Shepherd pre-trains a KG encoder via link prediction, then adds a downstream gene-ranking model. A key limitation is its reliance on an external candidate gene list, either expert-curated or generated by additional tools.

\section{Method} \label{sec:method}
\begin{figure}[t]
    \centering
    \includegraphics[width=1\textwidth]{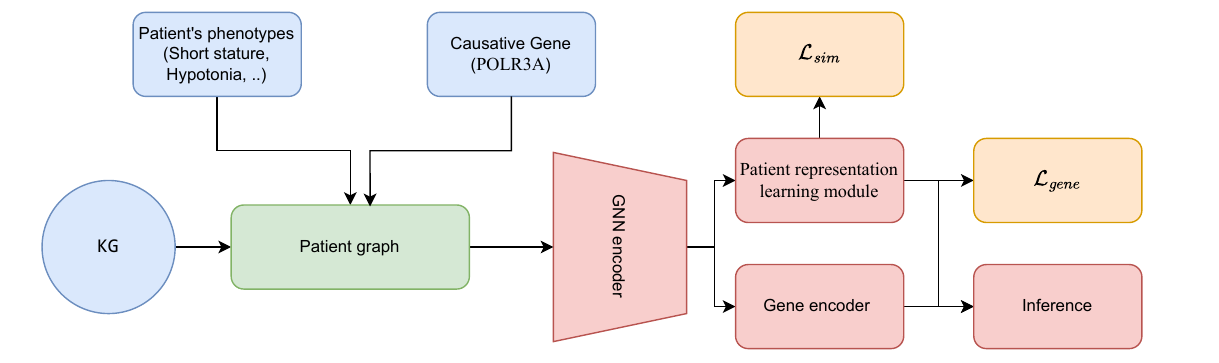}
    \caption{Overview of \methodName~for rare disease gene prioritization. The model constructs a patient-specific subgraph using phenotypes and a knowledge graph. GATv2 layers generate node embeddings, which are used to create patient and gene representations. These feed into loss functions to predict the most likely causative gene.}
    \label{fig:method}
\end{figure}

\usetikzlibrary{arrows, positioning, fit}
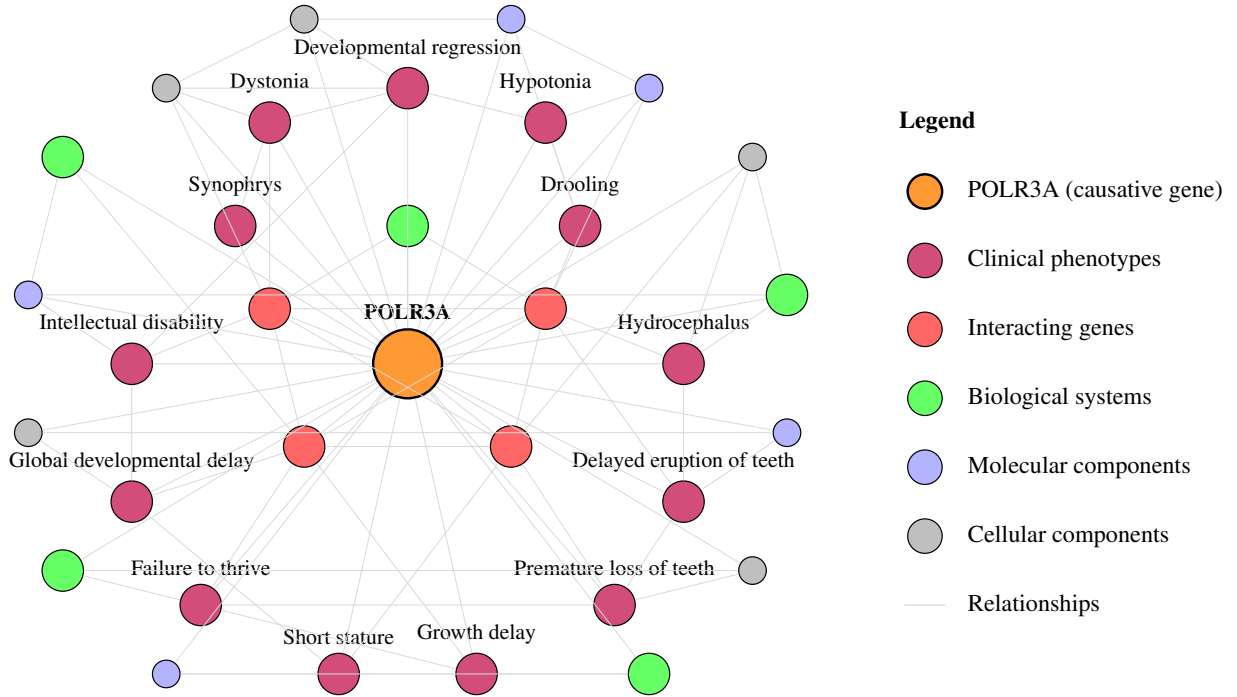
\begin{figure}[h]
\centering
\resizebox{1\textwidth}{!}{
\begin{tikzpicture}[
  center-node/.style={
    circle,
    fill=orange!80,
    draw=black,
    line width=1pt,
    minimum size=1cm,
    font=\footnotesize\bfseries
  },
  phenotype/.style={
    circle,
    fill=purple!70,
    draw=black,
    line width=0.5pt,
    minimum size=0.6cm,
    font=\tiny,
    label={[font=\footnotesize]above:#1}
  },
  gene/.style={
    circle,
    fill=red!60,
    draw=black,
    line width=0.5pt,
    minimum size=0.6cm
  },
  system/.style={
    circle,
    fill=green!60,
    draw=black,
    line width=0.5pt,
    minimum size=0.6cm
  },
  cellular/.style={
    circle,
    fill=gray!50,
    draw=black,
    line width=0.5pt,
    minimum size=0.4cm
  },
  component/.style={
    circle,
    fill=blue!30,
    draw=black,
    line width=0.5pt,
    minimum size=0.4cm
  },
  connection/.style={
    draw=gray!30,
    line width=0.3pt
  }
]

\node[center-node, label={[font=\footnotesize\bfseries]above:POLR3A}] (polr3a) at (0,0) {};

\node[gene] (gene1) at (-2,0.8) {};
\node[gene] (gene2) at (2,0.8) {};
\node[gene] (gene3) at (-1.5,-1.2) {};
\node[gene] (gene4) at (1.5,-1.2) {};

\node[phenotype={Dystonia}] (dystonia) at (-2,3.5) {};
\node[phenotype={Developmental regression}] (regression) at (0,4) {};
\node[phenotype={Hypotonia}] (hypotonia) at (2,3.5) {};
\node[phenotype={Hydrocephalus}] (hydro) at (4,0) {};
\node[phenotype={Delayed eruption of teeth}] (delayed) at (4,-2) {};
\node[phenotype={Premature loss of teeth}] (tooth_loss) at (3,-3.5) {};
\node[phenotype={Growth delay}] (growth) at (1,-4.5) {};
\node[phenotype={Short stature}] (short) at (-1,-4.5) {};
\node[phenotype={Failure to thrive}] (failure) at (-3,-3.5) {};
\node[phenotype={Global developmental delay}] (global) at (-4,-2) {};
\node[phenotype={Intellectual disability}] (intellectual) at (-4,0) {};
\node[phenotype={Synophrys}] (synophrys) at (-2.5,2) {};
\node[phenotype={Drooling}] (drooling) at (2.5,2) {};

\foreach \i/\type/\x/\y in {1/cellular/-3.5/4, 2/component/3.5/4, 3/system/-5/3, 4/cellular/5/3,
                            5/component/-5.5/1, 6/system/5.5/1, 7/cellular/-5.5/-1, 8/component/5.5/-1,
                            9/system/-5/-3, 10/cellular/5/-3, 11/component/-3.5/-4.5, 12/system/3.5/-4.5,
                            13/cellular/-1.5/5, 14/component/1.5/5, 15/system/0/2}
{
  \node[\type] (extra\i) at (\x,\y) {};
}

\foreach \node in {dystonia, regression, hypotonia, hydro, delayed, tooth_loss, growth, short, failure, global, intellectual, synophrys, drooling, gene1, gene2, gene3, gene4}
{
  \draw[connection] (polr3a) -- (\node);
}

\foreach \i in {1,...,15}
{
  \draw[connection] (polr3a) -- (extra\i);
}

\foreach \from/\to in {gene1/gene2, gene2/gene3, gene3/gene4, gene4/gene1, gene1/gene3, gene2/gene4}
{
  \draw[connection] (\from) -- (\to);
}

\foreach \from/\to in {dystonia/regression, regression/hypotonia, hypotonia/drooling,
                      hydro/delayed, delayed/tooth_loss, tooth_loss/failure, failure/growth,
                      growth/short, short/global, global/intellectual,
                      intellectual/regression, dystonia/synophrys, drooling/hypotonia}
{
  \draw[connection] (\from) -- (\to);
}

\foreach \from/\to in {gene1/dystonia, gene1/intellectual,
                      gene2/hydro, gene2/delayed, gene2/drooling,
                      gene3/growth, gene3/failure, gene3/global,
                      gene4/tooth_loss, gene4/short}
{
  \draw[connection] (\from) -- (\to);
}

\foreach \from/\to in {extra1/dystonia, extra1/regression, extra2/hypotonia,
                      extra4/hydro, extra5/intellectual,
                      extra6/hydro, extra7/global, extra8/delayed,
                      extra9/failure, extra10/tooth_loss, extra11/short,
                      extra12/growth, extra13/regression, extra14/hypotonia,
                      extra15/gene1, extra15/gene2}
{
  \draw[connection] (\from) -- (\to);
}

\foreach \from/\to in {extra1/gene1, extra2/gene2, extra3/gene3, extra4/gene4,
                      extra5/extra6, extra7/extra8, extra9/extra10,
                      extra11/extra12, extra13/extra14, extra1/extra13,
                      extra2/extra14, extra3/extra5, extra4/extra6}
{
  \draw[connection] (\from) -- (\to);
}

\node[anchor=west] at (7,3.5) {\textbf{Legend}};

\node[center-node, minimum size=0.5cm] (leg1) at (7.5,2.5) {};
\node[anchor=west] at (8,2.5) {POLR3A (causative gene)};

\node[phenotype={\phantom{x}}, minimum size=0.5cm] (leg2) at (7.5,1.5) {};
\node[anchor=west] at (8,1.5) {Clinical phenotypes};

\node[gene, minimum size=0.5cm] (leg3) at (7.5,0.5) {};
\node[anchor=west] at (8,0.5) {Interacting genes};

\node[system, minimum size=0.5cm] (leg4) at (7.5,-0.5) {};
\node[anchor=west] at (8,-0.5) {Biological systems};

\node[component, minimum size=0.5cm] (leg5) at (7.5,-1.5) {};
\node[anchor=west] at (8,-1.5) {Molecular components};

\node[cellular, minimum size=0.5cm] (leg6) at (7.5,-2.5) {};
\node[anchor=west] at (8,-2.5) {Cellular components};

\draw[connection, line width=0.5pt] (7.2,-3.5) -- (7.8,-3.5);
\node[anchor=west] at (8,-3.5) {Relationships};

\end{tikzpicture}
}
\caption{An example of a simplified patient-specific one-hop subgraph and its neighborhood, illustrating the relationships between the \textit{POLR3A} gene and associated genes, diseases, and phenotypes. The real patient subgraph maintains a similar structure but is substantially more complex.}
\label{fig:pg}
\end{figure}
We propose \methodName{}, a framework for predicting the causative gene in rare monogenic diseases by prioritizing genes associated with patient phenotypes. Our approach leverages the PrimeKG knowledge graph \(G\)~\citep{chandak2023building} to model gene-phenotype relationships. Specifically, \methodName{} learns informative representations from patient-specific subgraphs \(G_p\), which uniquely represent a patient based on their list of phenotypes and a set of candidate genes provided by clinicians or obtained from the knowledge graph. The patient subgraph is further augmented with additional information from the knowledge graph, such as related diseases and other relevant entities. As output, the model provides a ranked list of candidate genes, each associated with a relevance score indicating its likelihood of being causative for the patient. An overview of the proposed method is shown in \autoref{fig:method}.
\subsection{Problem Formulation}
Given a patient \(p\) with an associated set of phenotypes \(\mathcal{P}_p\), encoded using the Human Phenotype Ontology (HPO) system (e.g., \texttt{[HP:0004322, HP:0001263, \dots]}), and a complete candidate gene set \(\mathcal{G}\) (e.g., \texttt{[ENSG00000123066, ENSG00000141510, \dots]}), the set \(\mathcal{G}\) with the goal of identifying the causative gene \(g^* \in \mathcal{G}\) responsible for the patient’s condition (e.g., \(g^* = \texttt{ENSG00000165588}\), which corresponds to the \textit{OTX2} gene).

For each patient, the candidate gene list can be extracted from the global knowledge graph \( G \), either based on phenotype-gene associations or as specified by clinicians. The global knowledge graph \( G \) is represented as an undirected graph defined by \( G = (X, E, A) \), where \( X \in \mathbb{R}^{N \times d_{\text{in}}} \) denotes the set of node features, with \( N \) being the number of nodes in KG and \( d_{\text{in}} \) the feature dimension. Each node, corresponding to a gene, phenotype, or other biomedical entity, is initialized with a unique embedding obtained by pretraining \( G \) on a link prediction task, as described in \citep{alsentzer2022few}. The set \( E \) represents the edges capturing relationships between nodes, such as phenotype-gene associations and gene-gene interactions. The matrix \( A \) defines the connectivity structure of the graph and is described in detail in Section~\ref{sec:dataset}.
\subsection{Patient-specific Subgraphs}
\label{patient_subgraph}
To construct the patient-specific subgraph \( G_p \), we compute the shortest paths from each phenotype \( p_i \in \mathcal{P}_p \) to a provided or inferred list of candidate genes. The subgraph \( G_p \) includes all nodes along these paths, along with any additional nodes required to ensure connectivity. If no candidate gene list is available, the \( k \)-hop neighborhood can be extracted for each phenotype from \( G \), and all genes within this neighborhood are treated as candidates. Formally, \( G_p \) is represented as \( (X_p, E_p, A_p) \), where \( X_p \subseteq X \), \( E_p \subseteq E \), and \( A_p \subseteq A \). Here, \( X_p \in \mathbb{R}^{n \times d_{\text{in}}} \) denotes the node feature matrix, \( E_p \subseteq \{1, \dots, n\}^2 \) represents the edge set with the the \(n\) nodes in \(G_p\), and \( A_p \in \mathbb{R}^{|E_p| \times d_e} \) contains the edge attributes corresponding to \( E_p \), an example of the patient subgraph is illustrated \autoref{fig:pg}.
\subsubsection{Model}
\paragraph{GNN module} \label{par:gnn_module} For each \(G_p\), we apply a multi-layer graph convolutional network encoder to refine the original node embeddings. Let \( H^{(0)} = X_p \) denote the initial node features. Node representations are updated across \(L\) layers as follows:

\[
H^{(l)} = \text{Dropout}\left( \sigma\left( \text{LayerNorm}\left( \text{GATv2}(H^{(l-1)}, E, A) \right) \right) \right), \quad l = 1, \dots, L-1,
\]

where \( \sigma(\cdot) \) is a non-linear activation function, \( \text{LayerNorm}(\cdot) \) denotes layer normalization, and \( \text{Dropout}(\cdot) \) is a dropout operation applied for regularization during training. The final node representations are projected via a learnable linear map ($f_{\text{proj}}$):
\[
Z = f_{\text{proj}}\left( \text{GATv2}(H^{(L-1)}, E, A) \right).
\] The GATv2 is an attention graph operator from \citep{brody2022how}.

Let \( \{z_i\}_{i=1}^n \) be the node embeddings output by the GNN module. We partition these into phenotype and gene sets.
\paragraph{Patient representation learning module}\label{par:patient_module}

Phenotype embeddings are first projected by a multi-layer perceptron (MLP), denoted by $f_{\text{pheno}}$:
\[
\tilde{z}_i = f_{\text{pheno}}(\{z_i\}_{i \in \mathcal{P}_p}).
\]
To incorporate global context, \( m \) learnable memory vectors \( M \in \mathbb{R}^{m \times d} \) are concatenated with the phenotype embeddings and passed through multi-head self-attention:
\[
\hat{Z} = \text{MHA}([\tilde{Z}; M]),
\]
where MHA denotes standard multi-head attention as introduced in \citep{vaswani2017attention}.

The patient representation \(p \in \mathbb{R}^{d}\) is then obtained by masked mean pooling over phenotype nodes, followed by a two-layer MLP denoted by $f_{\text{patient}}$:
\[
p = f_{\text{patient}}\left(\frac{1}{|\mathcal{P}_p|} \sum_{i \in \mathcal{P}_p} \hat{Z}_i\right).
\]
\paragraph{Gene Encoder}\label{par:gene_module}
Gene node embeddings are processed via a Transformer-based encoder \citep{vaswani2017attention}, parameterized by $\theta$:
\[
\mathbf{G} = \theta(\{z_j\}_{j \in \mathcal{G}_p}),
\]
where \(\mathcal{G}_p\) are the gene indices in the patient graph and \(\mathbf{G} \in \mathbb{R}^{L \times d}\) are gene embeddings. 
\subsection{Losses}
\paragraph{Gene Loss}
To train the model, we adopt a contrastive loss framework. Let \(p \in \mathbb{R}^d\) be the patient embedding, \(\mathbf{G} = [g_1, \dots, g_L] \in \mathbb{R}^{L \times d}\) the candidate gene embeddings, and \(g^*\) the embedding of the causative gene. All embeddings are normalized to unit norm. We compute cosine similarities, where \(\tau\) is a learnable temperature parameter:
\[
\text{sim}(p, g_i) = \frac{\langle p, g_i \rangle}{\tau},
\]

Let \(\text{sim}^* = \text{sim}(p, g^*)\) be the similarity to the true gene. We apply semi-hard negative mining by first selecting a candidate set of negative genes satisfying:
\[
\text{sim}(p, g^-) < \text{sim}^* - \gamma,
\]
where \(\gamma\) is a margin hyperparameter. Among this candidate set, we select the negative gene with the highest similarity (i.e., the hardest semi-hard negative). If no semi-hard candidate exists, we select the negative gene with the highest overall similarity. This strategy balances learning from informative negatives while avoiding extreme outliers.

The loss is then computed using a margin-based triplet loss:
\[
\mathcal{L}_{\text{triplet}} = \max(0, \text{sim}(p, g^-) - \text{sim}(p, g^*) + \gamma).
\]

We further add an L2 norm regularization term to encourage both patient and gene embeddings to remain close to unit norm:
\[
\mathcal{L}_{\text{reg}} = \lambda \left| \|p\|_2 + \text{mean}\left(\|\mathbf{G}\|_2\right) - 2 \right|,
\]
where \(\lambda\) is a regularization weight and \(\|\mathbf{G}\|_2\) is the mean L2 norm across candidate gene embeddings.

The total loss is:
\[
\mathcal{L}_{\text{gene}} = \mathcal{L}_{\text{triplet}} + \mathcal{L}_{\text{reg}}.
\]

\paragraph{Patient Similarity Loss}
To encourage consistency across batches and improve generalization, we maintain a memory bank \(\mathcal{M}\) of past patient embeddings and their associated gene labels. Let \(\mathcal{M} = \{(p_i, g_i)\}\) be the set of embeddings in the memory bank.

The patient similarity loss is computed based on cosine similarities between patient embeddings having the same causative gene in the current and previous batches. We define:
\[
\mathcal{L}_{\text{sim}} = \mathcal{L}_{\text{within}} + \mathcal{L}_{\text{cross}},
\]
where \(\mathcal{L}_{\text{within}}\) compares embeddings within the current batch, and \(\mathcal{L}_{\text{cross}}\) compares them against the memory bank.

Each of \(\mathcal{L}_{\text{within}}\)  and \(\mathcal{L}_{\text{cross}}\) consists of two parts: pulling same-gene pairs together and pushing different-gene pairs apart.
\[
\mathcal{L}_{\text{pull}} = -\log \sigma\left(\frac{\text{sim}(p_i, p_j)}{\alpha}\right), \quad \text{for } g_i = g_j
\]
\[
\mathcal{L}_{\text{push}} = \max(0, \delta - (1 - \text{sim}(p_i, p_j))) \quad \text{for } g_i \ne g_j
\]

The final loss becomes:
\[
\mathcal{L}_{\text{total}} = \mathcal{L}_{\text{gene}} + \mathcal{L}_{\text{sim}},
\]
with \(\delta\) the margin and \(\alpha\) the temperature. The memory bank is updated per batch using a circular buffer strategy to retain a fixed number of embeddings.
\subsection{Inference}
At inference time, we construct the patient-specific subgraph \(G_p\) by extracting the \(k\)-hop neighborhood around each phenotype in \(\mathcal{P}_p\), thereby capturing the set of candidate genes \(\mathcal{G}_p\) associated with the patient's clinical profile (Section~\ref{patient_subgraph}). The model encodes \(G_p\) to produce a patient embedding \(p\) and gene embeddings \(\mathbf{G} = \{g_j\}_{j \in \mathcal{G}_p}\).

Relevance scores are computed via:
\[
\text{sim}(p, g_j) = \langle p, g_j \rangle, \quad \forall j \in \mathcal{G}_p.
\]

The final output is a ranked list of candidate genes:
\[
\mathcal{G}_p^{\text{ranked}} = \text{argsort}_{j \in \mathcal{G}_p} \left( \text{sim}(p, g_j) \right).
\]

\section{Experiments and Results}
\label{sec:impl_details}
\subsection{Dataset}
\label{sec:dataset}
We used two datasets in our experiments. The model was trained on a simulated dataset from \citep{alsentzer2023simulation}, which includes training (36,224 patients) and validation (6,080 patients) splits. Simulated data was chosen for its similarity to the Undiagnosed Diseases Network (UDN) dataset \citep{ramoni2017undiagnosed}, its larger size for training complex models, and its public availability, unlike most real datasets except MyGene2(\citep{mygene2}). A random subset of 320 patients from the original validation set was used as a final test set. Each simulated patient includes positive phenotypes and a challenging candidate gene list. Patients with causative genes missing from the knowledge graph were excluded from training. While our method theoretically can assess previously unseen genes, this requires the new gene to be manually connected to the existing KG (e.g., via related genes or diseases), which allows information and embeddings to propagate without the need to retrain the entire model, however, investigating the performance of it lies beyond the scope of the current work.

The second test dataset is the real-world MyGene2 set, originally comprising 146 patients and 48 unique causal genes, with an average of \(7.9 \pm 6.6\) phenotypes per patient. Data collection and preprocessing followed \citep{alsentzer2022few}. As MyGene2 lacks expert-curated candidate gene lists, we used it to test model performance without candidate genes. For efficiency, we restricted evaluation to the two-hop neighborhood around each patient; patients whose causative gene was unreachable within two hops were excluded, leaving 121 patients.  It is worth noting that the simulated dataset is similar in structure to the patients in the UDN \citep{ alsentzer2023simulation}, as it was generated based on and compared to it. However, it differs significantly from the MyGene2 dataset. Unlike MyGene2, the simulated dataset provides a list of candidate genes for each patient. Consequently, the patient subgraphs in the two datasets differ in both structure and size.

Our method uses a knowledge graph (KG) to augment patient information and identify candidate genes. We employed PrimeKG, developed in \citep{chandak2023building} and adapted for rare disease tasks in \citep{alsentzer2022few}, comprising 105,220 nodes and 1,095,469 edges. Nodes represent seven biological entity types: phenotypes (15,874), diseases (21,233), genes/proteins (21,610), pathways (2,516), molecular functions (11,169), cellular components (4,176), and biological processes (28,642), defined by biomedical vocabularies such as HPO \citep{kohler2019expansion} and ENSEMBL \citep{aken2016ensembl}, among others. Edges span 17 relation types, including protein-protein interactions (321,075), disease–phenotype (204,779 positive, 1,483 negative), phenotype–phenotype (21,925), phenotype–protein (10,518), and disease–protein associations (86,299), among others. Full KG details are in \citep{chandak2023building, alsentzer2022few}.

We use the terms \textit{gene} and \textit{protein} in this section interchangeably, as genes are represented by their encoded proteins in the knowledge graph. This reflects that most biological interaction data, such as protein-protein interactions and functional annotations, are defined at the protein level. Thus, gene references correspond to protein nodes acting as proxies for genes.

\subsection{Implementation Details}
We pre-trained the global graph \( G \) on a link prediction task \citep{alsentzer2022few} with 512-dimensional output embeddings. The GNN module used \( L = 3 \) GATv2 \citep{brody2022how} layers with \( h = 2 \) attention heads per layer, hidden dimensions \( d_{\text{hid}}^{(1)} = 1024 \), \( d_{\text{hid}}^{(2)} = 256 \), and output \( d_{\text{out}} = 512 \). LayerNorm \citep{ba2016layer} and LeakyReLU \citep{maas2013rectifier} were applied between layers, with dropout \( p = 0.4 \).  
Edge attributes (\( d_e = 15 \)) were incorporated into the attention mechanism to enhance relational modeling, and a final linear transformation was applied to project the resulting node representations into the output embedding space.

For patient representation learning, phenotype projection \( f_{\text{pheno}} \) used a two-layer MLP with ReLU \citep{krizhevsky2012imagenet} and LeakyReLU. The memory bank contained \( m = 128 \) learnable vectors (normal init., mean 0, std 1). Phenotype attention used 4 attention heads. Patient embeddings were aggregated and passed through a two-layer MLP with LeakyReLU. Gene embeddings were processed by a transformer encoder \citep{vaswani2017attention} with 4 layers, 8 attention heads, and an intermediate size of 2048.
Model training was performed using the AdamW optimizer \citep{loshchilov2017decoupled} with a learning rate of \(1 \times 10^{-4}\), combined with a cosine annealing learning rate scheduler \citep{loshchilov2016sgdr} configured with \( T_0 = 10 \) and a multiplier \( T_{\text{mult}} = 2 \). The patient neighbourhood was defined using \( k = 2 \) nearest neighbours. The following hyperparameters were used: margin \( \gamma = 0.3 \), contrastive temperature \( \tau = 0.12 \), regularization weight \( \lambda = 0.03 \), patient similarity temperature \( \alpha = 0.5 \), and similarity margin \( \delta = 0.8 \).
All models were implemented and trained using PyTorch, PyTorch Geometric, and Transformers. Encoder training ran for up to 135 epochs with early stopping after 25 epochs. 
\subsection{Evaluation Metrics}
We evaluated our model using Mean Reciprocal Rank (MRR) and normalized Discounted Cumulative Gain (nDCG) \citep{jarvelin2002cumulated}. MRR calculates the inverse rank of the first correct gene for each case and averages across cases, offering smooth sensitivity to ranking changes. nDCG accounts for the relevance of all ranked genes, applying a logarithmic penalty to lower ranks and normalizing against an ideal ranking, with scores ranging from 0 to 1. It places greater importance on highly ranked relevant genes, making it particularly informative for prioritization tasks. In contrast, \emph{hits@$j$} simply checks whether the correct gene appears within the top \(j\) positions, ignoring rankings beyond that threshold.
\subsection{Results and Discussions}
\begin{figure}[htbp]
\centering
\begin{minipage}[t]{0.49\textwidth}
\centering
\begin{tikzpicture}
\begin{axis}[
    width=1.0\textwidth,
    height=0.7\textwidth,
    xlabel={Top-$q$},
    ylabel={Percentage (\%)},
    title={MyGene2 Dataset},
    xtick={1,2,3,4,5,6,7},
    xticklabels={25,50,100,170,300,500,1000},
    grid=major,
    ymin=0, ymax=100,
    ]
    %
  \addplot[solid,mark=*,color=blue,mark size=1pt] coordinates {
    (1, 0.83) (2, 0.83) (3, 2.48) (4, 3.31) 
    (5, 11.57) (6, 14.05) (7, 19.83)
  };
  %
  \addplot[dashed,mark=square*,color=violet,mark size=1.5pt] coordinates {
    (1, 17.36) (2, 21.49) (3, 26.45) (4, 35.54) 
    (5, 49.59) (6, 54.55) (7, 71.07)
  };
  %
  \addplot[mark=diamond*,color=green!60!black,mark size=2pt] coordinates {
    (1, 23.14) (2, 30.58) (3, 37.19) (4, 44.63)
    (5, 52.07) (6, 55.37) (7, 69.42)
  };
\end{axis}
\end{tikzpicture}
\end{minipage}
\hfill
\begin{minipage}[t]{0.49\textwidth}
\centering
\begin{tikzpicture}
\begin{axis}[
    width=1.0\textwidth,
    height=0.7\textwidth,
    xlabel={Top-$q$},
    ylabel={Percentage (\%)},
    title={Test Dataset},
    xtick={1,2,3,4,5,6,7},
    xticklabels={25,50,100,170,300,500,1000},
    grid=major,
    ymin=0, ymax=100,
    ]
    %
  \addplot[solid,mark=*,color=blue,mark size=1pt] coordinates {
    (1, 8.44) (2, 12.50) (3, 17.81) (4, 25.62) 
    (5, 35.31) (6, 45.00) (7, 62.81)
  };
  %
  \addplot[dashed,mark=square*,color=violet,mark size=1.5pt] coordinates {
    (1, 56.56) (2, 68.44) (3, 76.25) (4, 82.19) 
    (5, 85.94) (6, 89.69) (7, 93.75)
  };
  %
  \addplot[mark=diamond*,color=green!60!black,mark size=2pt] coordinates {
    (1, 59.69) (2, 67.81) (3, 76.56) (4, 84.38)
    (5, 90.00) (6, 91.25) (7, 94.06)
  };
\end{axis}
\end{tikzpicture}
\end{minipage}
\caption{Match percentage comparison for MyGene2 (left) and simulated test (right) datasets, across top-$q$ values. Blue solid line with circles: \methodName \xspace with \(\mathcal{L}_{\text{gene}}\); violet dashed line with squares: with combined \(\mathcal{L}\) loss ; green line with diamonds: model with \(\mathcal{L}_{\text{sim}}\).}
\label{fig:match_percentage_charts}
\end{figure}
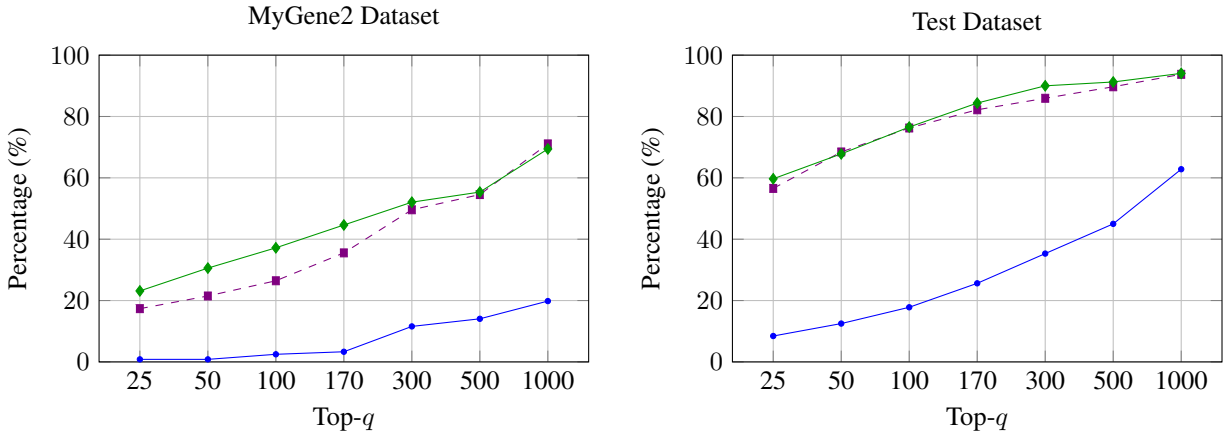

\begin{table}[tb]
\centering
\caption{Evaluation results using MRR and nDCG@K (in \%) are reported across different model groups for the simulated test dataset compared to SHEPHERD \citep{alsentzer2022few}. Higher values indicate better performance. The \textit{Pretrained embeddings} group corresponds to training initialized with embeddings pretrained on the knowledge graph for a link prediction task. The \textit{No embeddings} group refers to training from scratch. The \textit{Only patient phenotypes} group includes models trained solely on original patient phenotype data. Rows within each group represent different training setups using various loss function combinations.}
\resizebox{\linewidth}{!}{%
\begin{tabular}{llcccccc}
\toprule
\multicolumn{2}{l}{\textbf{Model}} & \textbf{MRR (\%)} $\uparrow$ & \textbf{nDCG@1} $\uparrow$ & \textbf{nDCG@3} $\uparrow$ & \textbf{nDCG@5} $\uparrow$ & \textbf{nDCG@10} $\uparrow$ & \textbf{nDCG@25} $\uparrow$ \\
\midrule
\multicolumn{8}{l}{\textbf{Competitive Method}} \\
\addlinespace
\multicolumn{2}{l}{SHEPHERD} & 79.35\xpm{0.00} & 66.88\xpm{0.00} & 80.07\xpm{0.00} & 82.77\xpm{0.00} & 84.26\xpm{0.00} & 84.43\xpm{0.00} \\
\midrule
\multicolumn{8}{l}{\textbf{PhenoKG (Ours)}} \\
\multicolumn{2}{l}{$\mathcal{L}_{\text{gene}}$\hspace{1.0ex}$\mathcal{L}_{\text{sim}}$
} & & & & & & \\
\addlinespace
\multicolumn{8}{l}{\textit{No Embeddings}} \\
\addlinespace
\cmark & \xmark & 85.32\xpm{1.52} & 76.67\xpm{2.70} & 86.55\xpm{1.15} & 87.73\xpm{1.13} & 88.91\xpm{1.18} & 88.94\xpm{1.14} \\
\xmark & \cmark & 86.81\xpm{0.84} & 79.69\xpm{0.68} & 87.54\xpm{1.10} & 88.91\xpm{0.90} & 89.69\xpm{0.78} & 90.01\xpm{0.66} \\
\cmark & \cmark & 86.60\xpm{1.36} & 79.27\xpm{1.50} & 87.45\xpm{1.61} & 88.73\xpm{1.31} & 89.41\xpm{1.16} & 89.84\xpm{1.06} \\
\addlinespace
\multicolumn{8}{l}{\textit{Only Patient Phenotypes}} \\
\addlinespace
\cmark & \xmark & 89.54\xpm{0.45} & 83.85\xpm{0.78} & 90.00\xpm{0.26} & 91.08\xpm{0.28} & 91.86\xpm{0.27} & 92.07\xpm{0.33} \\
\xmark & \cmark & 91.08\xpm{0.71} & 86.98\xpm{0.74} & 91.26\xpm{1.01} & 92.22\xpm{0.68} & 92.88\xpm{0.49} & 93.20\xpm{0.56} \\
\cmark & \cmark & 90.37\xpm{0.92} & 85.10\xpm{0.90} & 90.95\xpm{1.15} & 91.73\xpm{0.73} & 92.38\xpm{0.82} & 92.70\xpm{0.72} \\
\addlinespace
\multicolumn{8}{l}{\textit{Pretrained Embeddings}} \\
\addlinespace
\cmark & \xmark & 89.81\xpm{0.77} & 84.27\xpm{1.54} & 90.23\xpm{0.60} & 91.23\xpm{0.47} & 92.10\xpm{0.53} & 92.28\xpm{0.57} \\
\xmark & \cmark & 91.61\xpm{1.05} & 87.19\xpm{1.42} & 92.03\xpm{1.10} & 92.76\xpm{0.88} & 93.32\xpm{0.81} & 93.63\xpm{0.80} \\
\cmark & \cmark & 89.15\xpm{0.92} & 83.96\xpm{1.45} & 89.20\xpm{0.91} & 90.58\xpm{0.81} & 91.36\xpm{0.57} & 91.74\xpm{0.69} \\
\bottomrule
\end{tabular}
}
\label{tab:sim_grouped_ndcg_inputs}
\end{table}
\begin{table}[tb]
\centering
\caption{Evaluation results using MRR and nDCG@K (in \%) are reported across different model groups for the MyGene2 dataset. Higher values indicate better performance. The \textit{Pretrained embeddings} group corresponds to training initialized with embeddings pretrained on the knowledge graph for a link prediction task. The \textit{No embeddings} group refers to training from scratch. The \textit{Only patient phenotypes} group includes models trained solely on original patient phenotype data. Rows within each group represent different training setups using various loss function combinations.}
\resizebox{\linewidth}{!}{%
\begin{tabular}{llccccccccc}
\toprule
\multicolumn{2}{l}{\textbf{Model}} & \textbf{MRR} $\uparrow$ & \textbf{nDCG@1} $\uparrow$ & \textbf{nDCG@3} $\uparrow$ & \textbf{nDCG@5} $\uparrow$ & \textbf{nDCG@10} $\uparrow$ & \textbf{nDCG@25} $\uparrow$ & \textbf{nDCG@50} & \textbf{nDCG@75} & \textbf{nDCG@100} \\
\midrule
\multicolumn{11}{l}{\textbf{Competitive Method}} \\
\addlinespace
\multicolumn{2}{l}{SHEPHERD} & 19.02\xpm{0.00} & 11.57\xpm{0.00} & 15.42\xpm{0.00} & 17.37\xpm{0.00} & 19.28\xpm{0.00} & 27.45\xpm{0.00} & 29.21\xpm{0.00} & 30.03\xpm{0.00} & 30.54\xpm{0.00} \\
\midrule
\multicolumn{11}{l}{\textbf{PhenoKG (Ours)}} \\
\multicolumn{2}{l}{$\mathcal{L}_{\text{gene}}$\hspace{1.0ex}$\mathcal{L}_{\text{sim}}$
} & & & & & & \\
\addlinespace
\multicolumn{11}{l}{\textit{No Embeddings}} \\
\addlinespace
\cmark & \xmark & 5.74\xpm{0.99} & 1.65\xpm{0.67} & 2.69\xpm{1.19} & 3.95\xpm{1.12} & 5.74\xpm{0.87} & 9.39\xpm{1.27} & 12.77\xpm{1.62} & 14.62\xpm{2.19} & 16.19\xpm{2.09} \\
\xmark & \cmark & 7.27\xpm{2.47} & 2.75\xpm{0.78} & 5.15\xpm{2.40} & 6.28\xpm{3.25} & 7.87\xpm{3.89} & 10.36\xpm{3.56} & 12.53\xpm{2.75} & 14.47\xpm{2.18} & 15.54\xpm{2.31} \\
\cmark & \cmark & 5.87\xpm{0.94} & 2.48\xpm{0.67} & 3.76\xpm{0.46} & 4.32\xpm{1.07} & 5.87\xpm{0.90} & 7.97\xpm{1.76} & 11.69\xpm{1.77} & 13.49\xpm{1.92} & 14.59\xpm{1.63} \\
\addlinespace
\multicolumn{11}{l}{\textit{Only Patient Phenotypes}} \\
\addlinespace
\cmark & \xmark & 21.52\xpm{1.76} & 11.02\xpm{2.81} & 20.00\xpm{1.39} & 21.72\xpm{1.63} & 25.30\xpm{1.53} & 28.38\xpm{1.27} & 29.48\xpm{1.38} & 30.22\xpm{1.10} & 30.78\xpm{1.54} \\
\xmark & \cmark & 19.19\xpm{0.88} & 7.16\xpm{0.39} & 16.55\xpm{0.96} & 20.60\xpm{0.87} & 25.07\xpm{1.78} & 27.13\xpm{0.59} & 28.81\xpm{0.64} & 29.73\xpm{0.81} & 30.28\xpm{0.79} \\
\cmark & \cmark & 19.49\xpm{1.56} & 8.54\xpm{2.55} & 16.55\xpm{0.62} & 20.02\xpm{2.12} & 24.28\xpm{1.38} & 27.18\xpm{1.34} & 29.43\xpm{1.11} & 30.08\xpm{1.04} & 30.59\xpm{0.95} \\
\addlinespace
\multicolumn{11}{l}{\textit{Pretrained Embeddings}} \\
\addlinespace
\cmark & \xmark & 23.40\xpm{1.43} & 13.22\xpm{1.17} & 21.71\xpm{0.72} & 25.02\xpm{1.85} & 27.10\xpm{3.05} & 29.68\xpm{2.49} & 30.86\xpm{1.81} & 31.55\xpm{1.52} & 32.41\xpm{1.67} \\
\xmark & \cmark & 21.13\xpm{2.63} & 10.19\xpm{2.55} & 18.10\xpm{3.42} & 22.60\xpm{3.42} & 25.52\xpm{2.91} & 28.83\xpm{2.27} & 30.58\xpm{1.81} & 31.15\xpm{1.68} & 31.53\xpm{1.77} \\
\cmark & \cmark & 24.64\xpm{4.57} & 15.15\xpm{3.33} & 22.74\xpm{5.28} & 25.33\xpm{5.40} & 28.17\xpm{4.73} & 30.86\xpm{4.40} & 32.30\xpm{4.42} & 33.04\xpm{4.32} & 33.64\xpm{4.64} \\
\bottomrule
\end{tabular}
}
\label{tab:mygene2_grouped_ndcg_inputs}
\end{table}
The proposed model and the competitive Shepherd model~\citep{alsentzer2022few} were evaluated under the same experimental setup, data splits, and preprocessing protocols. Shepherd was selected for comparison due to its use of knowledge graphs and its superior performance over other models~\citep{alsentzer2022few}. We tested multiple configurations of our model using different combinations of loss functions: gene loss only, patient similarity loss only, and a combined loss. Additionally, we assessed variants using three embedding strategies: (1) \textit{Pretrained Embeddings}, where node embeddings were initialized from link prediction pretraining; (2) \textit{No Embeddings}, with randomly initialized embeddings; and (3) \textit{Only Patient Phenotypes}, where only real patient phenotype nodes were used for subgraph construction and the model used pretrained embeddings. The \textit{Pretrained Embeddings} and \textit{No Embeddings} configurations included all phenotypes in the patient graph. Each configuration was trained three times, and we report the mean performance along with the standard deviation.

We also evaluated causative gene identification via patient similarity: each test patient embedding was compared to training and validation embeddings, with top-$q$ matches assessed for shared causative genes (\autoref{fig:match_percentage_charts}).
It is important to note that the MyGene2 and simulated test datasets differ in graph structure. MyGene2 includes portions of the KG not seen during training. Pretraining was shown to be not critical when training data covered the full KG or for deployment on similar datasets (\autoref{tab:sim_grouped_ndcg_inputs}), but was necessary for generalization to novel phenotype-gene combinations in MyGene2 (\autoref{tab:mygene2_grouped_ndcg_inputs}). On the simulated dataset, the \textit{Only Patient Phenotypes} model trained with similarity loss achieved an MRR of \(91.08 \pm 0.71\%\) and nDCG@1 of \(86.98 \pm 0.74\%\), outperforming Shepherd (MRR \(79.35\%\), nDCG@1 \(66.88\%\)). On MyGene2, the \textit{Pretrained Embeddings} model with combined loss obtained the best performance (MRR \(24.64 \pm 4.57\%\), nDCG@1 \(15.15 \pm 3.33\%\)), compared to Shepherd (MRR \(19.02\%\), nDCG@1 \(11.57\%\)).

We compared three loss function configurations. Models trained solely with gene loss performed well on the causative gene prediction task but poorly on the auxiliary patient similarity task (\autoref{fig:match_percentage_charts}). Models trained solely with patient similarity loss excelled at patient similarity-based gene identification and delivered competitive results on causative gene prediction, with only a minor drop in top-nDCG performance. The observed behavior can be attributed to the absence of alignment between patient embeddings across batches, which led to divergent representations even among patients sharing the same causative gene. Each patient’s unique graph structure causes patient embeddings to capture both shared and distinct attributes.

Models trained with the combined loss delivered the most robust and consistent performance on the MyGene2 dataset by enforcing alignment across patient representations while preserving individuality. This observation suggests an interesting direction for future work: understanding how patient embeddings relate when causative genes differ. If a model trained solely with gene loss yields patient representations that still cluster meaningfully, what other aspects of patient phenotype space are being captured? As our quantitative results suggest, patient embeddings encode more than just gene identity, which warrants further exploration.

Finally, analysis in \autoref{fig:match_percentage_charts} highlights substantial distributional differences between the MyGene2 and simulated datasets. When patients in the test set closely resemble training patients, prediction accuracy aligns across datasets. However, mismatches in patient phenotype distributions explain the performance gap observed between the datasets.

\section{Conclusion}
In this work, we proposed PhenoKG, a method that operates solely on patient phenotype data to produce a ranked list of potential causative genes, enabling the identification of novel or previously unobserved phenotype–gene associations not explicitly represented in the knowledge graph. The method does not rely on genetic data or predefined candidate lists and outperforms existing approaches across multiple gene prioritization metrics. It can be used as a pre-filter to narrow the search space before variant analysis and, following this pre-filtering, as a gene prioritization tool to rank the remaining candidate genes. However, while it exceeds recent state-of-the-art methods without candidate genes, its performance remains insufficient for standalone clinical application and is better suited as part of a multi-step filtering and prioritization pipeline.

\bibliographystyle{plainnat}
\bibliography{ref}

\appendix

\end{document}